\documentclass[10pt, a4paper]{article}
\usepackage{lrec}
\usepackage{multibib}
\newcites{languageresource}{Language Resources}
\usepackage{graphicx}
\usepackage{tabularx}
\usepackage{soul}
\usepackage{xcolor}
\usepackage{booktabs}
\usepackage{latexsym}
\usepackage{url}
\usepackage{enumitem} 
\usepackage{enumerate}
\usepackage[normalem]{ulem}  

\usepackage{epstopdf}
\usepackage[latin1]{inputenc}

\usepackage{hyperref}
\usepackage{xstring}

  \title{Word Affect Intensities}

\name{Saif M. Mohammad}

\address{National Research Council Canada\\
   {\tt saif.mohammad@nrc-cnrc.gc.ca}\\
       }

\abstract{
Words often convey affect---emotions, feelings, and attitudes. Further, different words can convey affect to various degrees (intensities).
However, existing manually created lexicons for basic emotions (such as anger and fear)  indicate only coarse categories of affect association (for example, associated with anger or not associated with anger). Automatic lexicons of affect provide fine degrees of association, but they tend not to be accurate as human-created lexicons.
Here, for the first time, we present a manually created affect intensity lexicon with real-valued scores of intensity for four basic emotions: anger, fear, joy, and sadness. 
(We will subsequently add entries for more emotions such as disgust, anticipation, trust, and surprise.)
We refer to this dataset as the {\it NRC Affect Intensity Lexicon}, or {\it AIL} for short. AIL has entries for close to 6,000 English words.  
We used a technique called best--worst scaling (BWS) to create the lexicon.
BWS improves annotation consistency and obtains reliable fine-grained scores (split-half reliability $> 0.91$).
We also compare the entries in AIL with the entries in the {\it NRC VAD Lexicon}, which has valence, arousal, and dominance (VAD) scores for 20K English words. We find that anger, fear, and sadness words, on average, have very similar VAD scores. However, sadness words tend to have slightly lower dominance scores than fear and anger words.  
The Affect Intensity Lexicon has applications in automatic emotion analysis in a number of domains such as commerce, education, intelligence, and public health. AIL is also useful in the building of natural language generation systems. 
\newline \Keywords{emotion intensity, emotion lexicon, emotion analysis, crowdsourcing, best--worst scaling, sentiment analysis} }

\begin{document}

\maketitleabstract

\section{Introduction}
 \setitemize[0]{leftmargin=*}
 \setenumerate[0]{leftmargin=*}

\noindent Words often convey affect---emotions, feelings, and attitudes. Some words have affect as a core part of their meaning.
For example, {\it dejected} and {\it wistful} denotate some amount of sadness (and are thus associated with sadness).
On the other hand, some words are associated with affect even though they do not denotate affect. For example,
{\it failure} and {\it death} describe concepts that are usually accompanied by sadness and thus they connotate some amount of sadness.
Lexicons of word--affect association have numerous applications, including: tracking
brand and product perception, tracking support for issues and policies, tracking public health and well-being,
literary analysis, and developing more natural dialogue systems. 
Past work on manually compiling affect lexicons has focused on denotative words \cite{Wiebe05,FranciscoG06,StrapparavaV04}.
A notable exception to this is the NRC Emotion Lexicon, which includes words that are associated with (or connotate) an emotion \cite{MohammadT13}.

Words can be associated with different intensities (or degrees) of an emotion. For example, most people will agree that the word {\it outrage} is associated with a greater degree of anger (or more anger) than the word {\it irritate}.
However, existing manually created affect lexicons for basic emotions such as anger and fear
do not provide scores for 
the intensity of the emotion. 
Annotating instances for fine-grained intensity of affect is a substantially more difficult undertaking than categorical annotation.
It is particularly hard to ensure consistency
(both across responses by different annotators and within the responses produced by the same annotator). 

{\it Best--Worst Scaling (BWS)} 
is an annotation scheme that  addresses these limitations
by employing comparative annotations
\cite{Louviere_1991,Cohen_2003,Louviere2015,KiritchenkoM2017bwsvsrs}.
Annotators are given $n$ items at a time (an $n$-tuple, where $n > 1$ and commonly $n= 4$). They are asked which item is the {\it best} (highest in terms of the property of interest) and which is the {\it worst} (least in terms of the property of interest).
When working on $4$-tuples, best--worst annotations are particularly efficient because each best and worst annotation will reveal the order of
five of the six item pairs (i.e., for a 4-tuple with items A, B, C, and D, if A is the best, and D is the worst, then A $>$ B, A $>$ C, A $>$ D, B $>$ D, and C $>$ D).

We can calculate real-valued scores of association between the items and the property of interest from the best--worst annotations for a set of $4$-tuples \cite{Orme_2009,flynn2014}. 
The scores can be used to rank items by the degree of association with the property of interest.
It has been empirically shown that three annotations each for $2N$ $4$-tuples is sufficient for obtaining reliable scores (where $N$ is the number of items) \cite{Louviere_1991,maxdiff-naacl2016}.\footnote{At its limit, when $n=2$, BWS becomes a {\it paired comparison} \cite{thurstone1927law,david1963method}, but then a much larger set of tuples need to be annotated (closer to $N^2$).}
Kiritchenko and Mohammad \shortcite{KiritchenkoM2017bwsvsrs} showed through empirical experiments that BWS produces more reliable and more discriminating scores than those obtained using rating scales.

Here, for the first time, we create an affect intensity lexicon with real-valued scores of association for four basic emotions (anger, fear, joy, and sadness) using best--worst scaling.
For a given word and emotion X, the scores range from 0 to 1. 
A score of 1 means that the word conveys the highest intensity (amount) of emotion X.
A score of 0 means that the word conveys the lowest intensity (amount) of emotion X.
We will refer to this lexicon as the {\it NRC Affect Intensity Lexicon (AIL)}.
AIL includes entries for close to 6,000 English words. 
It includes common English terms as well as terms that are more prominent in social media platforms, such as Twitter.
It includes terms that are associated with emotions to various degrees.
For a given emotion, this even includes some terms that may not predominantly convey that emotion (or that convey an antonymous emotion), and yet tend to co-occur with terms that do.
Antonymous terms tend to co-occur with each other more often than chance, and are particularly problematic when one uses automatic
co-occurrence-based statistical methods to capture word--emotion connotations.
Thus, it is  particularly beneficial to have manual annotations of affect intensity for these terms.

We show that repeat annotations of the terms in the Affect Intensity Lexicon with independent annotators lead to affect association scores that are close to the scores obtained originally (Spearman Rank correlations of  0.92; Pearson correlation: 0.91).
The fine-grained scores obtained with BWS  and the high correlations on repeat annotations indicate that BWS is both markedly discriminative (helps identify small differences in affect intensity) and markedly reliable (provides stable outcomes).


We also compare the entries in AIL with the entries in the {\it NRC VAD Lexicon}, which has valence, arousal, and dominance (VAD) scores for 20K English words. We find that anger, fear, and sadness words, on average, have very similar VAD scores. However, sadness words tend to have slightly lower dominance scores than fear and anger words.  
The Affect Intensity Lexicon has applications in automatic emotion analysis in a number of domains such as commerce, education, intelligence, and public health. AIL is also useful in the building of natural language generation systems. We have made the NRC  Affect Intensity Lexicon freely available for, non-commercial, research purposes.\footnote{www.saifmohammad.com/WebPages/AffectIntensity.htm}

We begin with a brief overview of the related work (Section 2), followed by a description of how we created the NRC Affect Intensity lexicon (Section 3). 
In Section 4, we study the valence, arousal, and dominance scores of words in the Affect Intensity Lexicon.
In Section 5, we present experiments on the reliability of the annotations. 
In Section 6, we outline various applications of the NRC Affect Intensity lexicon. Finally, in Section 6, we present concluding remarks.

\section{Related Work}

\noindent  Psychologists have argued that  some emotions are more basic than others  \cite{Ekman92,Plutchik80,frijda1988laws,Parrot01}.\footnote{However, 
they disagree on which emotions (and how many) should be classified as basic emotions---some propose 6, some 8, some 20, and so on.}
Thus, most work on capturing word--emotion associations has focused 
on a handful of emotions, especially since
manually annotating for a large number of emotions is arduous. 
In this project, we focus on four emotions common among the many proposals for basic emotions \cite{Plutchik80,Ekman92,Parrot01}: anger, fear, joy, and sadness. 

There is a large body of work on creating valence or sentiment lexicons, including the General Inquirer \cite{Stone66}, ANEW \cite{nielsen2011new,bradley1999affective},
MPQA \cite{Wiebe05}, NRC VAD Lexicon by \cite{Mohammad-NRC-Norms},
and the lexicon by \newcite{warriner2013norms}.
The work on creating lexicons for categorical emotions such as joy, sadness, fear, etc, is comparatively small.
WordNet Affect Lexicon \cite{StrapparavaV04} has a few hundred words annotated with the emotions they evoke.\footnote{http://wndomains.fbk.eu/wnaffect.html}
It was created by manually identifying the emotions of a few seed words and then marking all
their WordNet synonyms as having the same emotion. 
The NRC Emotion Lexicon was created by crowdsourcing and it includes entries for
about 14,000 words and eight Plutchik emotions 
\cite{MohammadT13,MohammadT10}.\footnote{http://www.purl.org/net/saif.mohammad/research}
It also includes entries for positive and negative sentiment.

Most prior work in sentiment analysis describes machine learning systems trained and tested on data with coarse categorical annotations.
 This is not surprising, because it is difficult for humans to directly provide valence (sentiment) scores at a fine granularity.   A common problem is inconsistencies in annotations among different annotators. 
One annotator might assign a score of 7.9 to a word, whereas another annotator may assign a score of 6.2 to the same word. 
 It is also common that the same annotator assigns different scores to the same word 
 at different points in time.
 Further, annotators often have a bias towards different parts of the scale, known as {\it scale region bias}.
 Despite this,  a key question is whether humans are able to distinguish affect at only four or five coarse levels, or whether we can discriminate across much smaller affect intensity differences.

Best--Worst Scaling (BWS) was developed by \newcite{Louviere_1991}, building on some ground-breaking research in the 1960's in mathematical psychology and psychophysics by Anthony A. J. Marley and Duncan Luce.  However, it is not  well known outside the areas of choice modeling and marketing research.
Within the NLP community, BWS has thus far been used for creating datasets for relational similarity
\cite{jurgens-EtAl:2012:STARSEM-SEMEVAL}
and word-sense disambiguation \cite{Jurgens2013EmbracingAA}.
\newcite{Mohammad-NRC-Norms} used best--worst scaling to annotate about 20K words for valence, arousal, and dominance.
In this work, we use BWS to annotate
words for intensity (or degree) of basic emotions. With BWS we address the challenges of direct scoring, and
produce more reliable emotion intensity scores.  Further, this will be the first dataset that will also include
emotion scores for words common in social media.

There is growing work on automatically determining word--emotion associations
\cite{COIN:COIN12024,Mohammad12,StrapparavaV04}.
These automatic methods  often assign a real-valued score  representing the degree of association. Further, these association scores are likely to be somewhat correlated with the intensity of the emotion. The Affect Intensity Lexicon can be used to judge the quality of the automatic lexicons, and also to explore the extent of correlation between emotion association and emotion intensity.

\section{NRC Affect Intensity Lexicon}
\noindent  We now present how we created the NRC Affect Intensity Lexicon. The two sub-sections below describe how we chose the terms to be annotated and how we annotated the chosen terms, respectively.

\subsection{Term Selection}
\noindent  We chose to annotate commonly used English terms, as well as
terms common in social media texts,
so that the resulting lexicon can be applied widely.
Twitter has a large and diverse user base, which
entails rich 
textual content.\footnote{Twitter is an online social networking and microblogging service where users post and read messages that are 
 up to 140 characters long.  The posts are called tweets.}
Tweets have plenty of non-standard language such as
emoticons, emojis, creatively spelled words ({\it happee}), hashtags ({\it \#takingastand, \#lonely}) and conjoined words ({\it loveumom}).  
Tweets are often used to convey one's emotions, opinions towards products, and stance over
issues. Thus, emotion analysis of tweets is particularly compelling.
Therefore, apart from common English terms, we also chose to annotate terms common in tweets.

Since most words do not convey a particular emotion to a marked degree, annotating all words for all emotions
is sub-optimal. Thus, for each of the eight emotions, we created separate lists of terms that satisfied either one of the two properties listed below:
\begin{itemize}
\item The word is already known to be associated with the emotion (although the intensity of emotion it conveys is unknown).
\item The word has a tendency to occur in tweets that express the emotion.
\end{itemize}
\noindent With these properties in mind, for our annotation, we included terms from two separate sources:
\begin{itemize}
\item The words listed in the NRC Emotion Lexicon that are marked as being associated with any of the Plutchik emotions.
\item The words that tend to co-occur more often than chance with emotion-word hashtags in a large tweets corpus. (Emotion-word hashtags, such as {\it \#angry, \#fear,} and {\it \#happiness}, act as noisy labels of the corresponding emotions.)
\end{itemize}
\noindent 
Since the NRC Emotion Lexicon \cite{MohammadT13,MohammadT10} includes only those terms that occur frequently in the Google n-gram corpus \cite{BrantsF06}, these terms satisfy the `commonly used terms' criterion as well.

The Hashtag Emotion Corpus \cite{Mohammad12} has tweets that each have at least one emotion-word hashtag. The emotion-word hashtags 
corresponding to the eight basic Plutchik emotions. As mentioned before, we consider the emotion-word hashtags as (noisy) labels of the corresponding emotions.
For every word 
that occurred more than ten times in the corpus, we computed the pointwise mutual information (PMI) between the word and each of the emotion labels. 
If a word has a greater-than-chance tendency to occur in tweets with a particular emotion label, then
it will have a PMI score that is greater than 0.
For each emotion, 
we included all terms in the Hashtag Emotion Corpus \cite{Mohammad12} that had a PMI $> 1$.
Note that this set of terms included both terms that are more common in social media communication (for example, {\it soannoyed, grrrrr, stfu}, and {\it thx})
as well as regular English words.\footnote{Some of the terms included from tweets were deliberate spelling variations of English words, for example, {\it bluddy} and {\it sux}.}

\begin{table*}[t!]
\begin{center}
\small{
\begin{tabular}{lrrrrrrrr}
\hline 
			 	& 	&\bf Location of 	&\bf Annotation & &  & &   	&\bf \#Best--Worst\\
{\bf Dataset} 	& \bf \#words	&\bf Annotators &\bf Item  & \bf \#Items & \bf \#Annotators &\bf MAI  		&\bf \#Q/Item	&\bf Annotations \\\hline
anger 		& 1,483 &USA  & 4-tuple of words	 & 2,966 & 119 & 4 & 2	& 12,212\\
fear 		& 1,765 &USA & 4-tuple of words	& 3,530 & 82	& 4 & 2	& 14,129 \\
joy   		& 1,268 & USA & 4-tuple of words	& 2,536 & 76 & 4 & 2 & 10,365	\\
sadness 		& 1,298 & USA & 4-tuple of words	& 2,596 & 76 & 4	&2 & 10,429\\
\hline
\bf Total  & \bf 5,814		& & 	& & & & & \bf 47,135\\
 \hline
\end{tabular}
}
\vspace*{-2mm}
\caption{\label{tab:ann} {Summary details of the current annotations done for the NRC Affect Intensity Lexicon. MAI = median number of annotations per item. Q = questions.}
}
\end{center}
\end{table*}

\subsection{Annotating for Affect Intensity with Best--Worst Scaling}

\noindent  For each emotion, the annotators were presented with four words at a time (4-tuples) and asked to select the word that
conveys the highest 
emotion intensity and the word that
conveys the lowest emotion intensity.
$2\times N$ (where $N$ is the number of words to be annotated) distinct 4-tuples were randomly generated in such a manner that
each word is seen in eight different 4-tuples, and no two 4-tuples had more than two items in common. We used the script provided by \newcite{maxdiff-naacl2016} to obtain the 4-tuples to be annotated.\footnote{http://saifmohammad.com/WebPages/BestWorst.html}
A sample questionnaire is shown below. 


{
\noindent\makebox[\linewidth]{\rule{0.5\textwidth}{0.4pt}}
{ 
\noindent {\bf Words Associated With Most And Least Anger}\\[-5pt]

\noindent Words can be associated with different degrees of an emotion. For example, most people will agree that the word {\it condemn} is associated with a greater degree of anger (or more anger) than the word {\it irritate}. The goal of this task is to determine the degrees of anger associated with words. Since it is hard to give a numerical score indicating the degree of anger, we will give you four different words and ask you to indicate to us:\\[-12pt]
\begin{itemize}
\item the word that is associated with the MOST anger\\[-15pt] 
\item the word that is associated with the LEAST anger\\[-12pt]
\end{itemize}
\noindent A rule of thumb that may be helpful is that a word associated with more anger tends to occur in many angry sentences, whereas a word associated with less anger tends to occur in fewer angry sentences.\\ [-4pt]



\noindent {\bf Important Notes}\\[-13pt]
\begin{itemize}
\item Some words, such as {\it furious} and {\it irritated}, are not only associated with anger, they also explicitly express anger. Others do not express anger, but they are associated with the emotion; for example, argument and corruption are associated with anger.  To be selected as `associated with MOST anger' or `associated with LEAST anger', a word does not have to explicitly express anger.\\[-13pt]
\item Some words have more than one meaning, and the different meanings may be associated with different degrees of anger. If one of the meanings of the word is strongly associated with anger, then base your response on that meaning of the word.\\ 
\end{itemize}

\noindent {\bf EXAMPLE}\\[-9pt]

\noindent Q1. Identify the term associated with the MOST anger.\\[-18pt]
 \begin{itemize}
 \item tree\\[-20pt]
 \item grrr\\[-20pt]
 \item boiling\\[-20pt]
 \item vexed\\[-18pt]
 \end{itemize}
\noindent Ans: boiling\\[-7pt]

\noindent Q2. Identify the term associated with the LEAST anger.\\[-18pt]
\begin{itemize}
 \item tree\\[-20pt]
 \item grrr\\[-20pt]
 \item boiling\\[-20pt]
 \item vexed\\[-18pt]
 \end{itemize}
\noindent Ans: tree\\[-16pt]

}
\noindent\makebox[\linewidth]{\rule{0.5\textwidth}{0.4pt}}

}
\noindent The questionnaires for other emotions are similar. 

\begin{table*}[t!]
\begin{center}
{\small
\begin{tabular}{lrrlrrlrrlr}
\hline
Word	&Anger	&	&Word	&Fear	&	&Word	&Joy	&	&Word	&Sadness\\\hline
{\it outraged}	&0.964	&	&{\it horror}	&0.923	&	&{\it sohappy}	&0.868	&	&{\it sad}	&0.844\\
{\it brutality}	&0.959	&	&{\it horrified}	&0.922	&	&{\it superb}	&0.864	&	&{\it suffering}	&0.844\\
{\it satanic}	&0.828	&	&{\it hellish}	&0.828	&	&{\it cheered}	&0.773	&	&{\it guilt}	&0.750\\
{\it hate}	&0.828	&	&{\it grenade}	&0.828	&	&{\it positivity}	&0.773	&	&{\it incest}	&0.750\\
{\it violence}	&0.742	&	&{\it strangle}	&0.750	&	&{\it merrychristmas}	&0.712	&	&{\it accursed}	&0.697\\
{\it molestation}	&0.742	&	&{\it tragedies}	&0.750	&	&{\it bestfeeling}	&0.712	&	&{\it widow}	&0.697\\
{\it volatility}	&0.687	&	&{\it anguish}	&0.703	&	&{\it complement}	&0.647	&	&{\it infertility}	&0.641\\
{\it eradication}	&0.685	&	&{\it grisly}	&0.703	&	&{\it affection}	&0.647	&	&{\it drown}	&0.641\\
{\it cheat}	&0.630	&	&{\it cutthroat}	&0.664	&	&{\it exalted}	&0.591	&	&{\it crumbling}	&0.594\\
{\it agitated}	&0.630	&	&{\it pandemic}	&0.664	&	&{\it woot}	&0.588	&	&{\it deportation}	&0.594\\
{\it defiant}	&0.578	&	&{\it smuggler}	&0.625	&	&{\it money}	&0.531	&	&{\it isolated}	&0.547\\
{\it coup}	&0.578	&	&{\it pestilence}	&0.625	&	&{\it rainbow}	&0.531	&	&{\it unkind}	&0.547\\
{\it overbearing}	&0.547	&	&{\it convict}	&0.594	&	&{\it health}	&0.493	&	&{\it chronic}	&0.500\\
{\it deceive}	&0.547	&	&{\it rot}	&0.594	&	&{\it liberty}	&0.486	&	&{\it injurious}	&0.500\\
{\it unleash}	&0.515	&	&{\it turbulence}	&0.562	&	&{\it present}	&0.441	&	&{\it memorials}	&0.453\\
{\it bile}	&0.515	&	&{\it grave}	&0.562	&	&{\it tender}	&0.441	&	&{\it surrender}	&0.453\\
{\it suspicious}	&0.484	&	&{\it failing}	&0.531	&	&{\it warms}	&0.391	&	&{\it beggar}	&0.422\\
{\it oust}	&0.484	&	&{\it stressed}	&0.531	&	&{\it gesture}	&0.387	&	&{\it difficulties}	&0.421\\
{\it ultimatum}	&0.439	&	&{\it disgusting}	&0.484	&	&{\it healing}	&0.328	&	&{\it perpetrator}	&0.359\\
{\it deleterious}	&0.438	&	&{\it hallucination}	&0.484	&	&{\it tribulation}	&0.328	&	&{\it hindering}	&0.359\\

\hline
\end{tabular}
\caption{\label{tab:examples} {Example entries for four emotions in the NRC Affect Intensity Lexicon. For each emotion, the table shows every 100th and 101st entry, when ordered by decreasing emotion intensity.}}
}
\end{center}
\end{table*}

We setup four crowdsourcing tasks corresponding to the four basic emotions.
The 4-tuples of words were uploaded for annotation on the crowdsourcing platform, CrowdFlower.\footnote{http://www.crowdflower.com}
We obtained annotations from native speakers of English residing in the United States of America. Annotators were free to provide responses to as many 4-tuples as they wished. 
The annotation tasks were approved by the National Research Council Canada's Institutional Review Board, which reviewed the proposed methods to ensure that they were ethical.

About 5\% of the data was annotated internally beforehand (by the author). These questions are referred to as gold questions. 
The gold questions are interspersed with other questions.
If one gets a gold question wrong, they are immediately notified of it. If one's accuracy on the gold questions falls below 70\%, they are refused further annotation, 
and all of their annotations are discarded. This serves as a mechanism to avoid malicious or random annotations.
 In addition, the gold questions serve as examples to guide the annotators.
 
\newcite{maxdiff-naacl2016} showed that using just three annotations per 4-tuple produces highly reliable results. 
 In task settings, we specified that we needed annotations from four people for each word.\footnote{Note that since each word occurs in eight different 4-tuples, each word is involved in $8 \times 4 = 32$ best--worst judgments.} However, because of the way the gold questions work in CrowdFlower, they were annotated by more than four people. Nonetheless, the median number of annotations is four (same as the minimum number of annotations). A total of 47,135 pairs of responses (best and worst) were obtained (see Table~\ref{tab:ann}).

\begin{figure}[t]
\centering
 \includegraphics[width=3.2in]{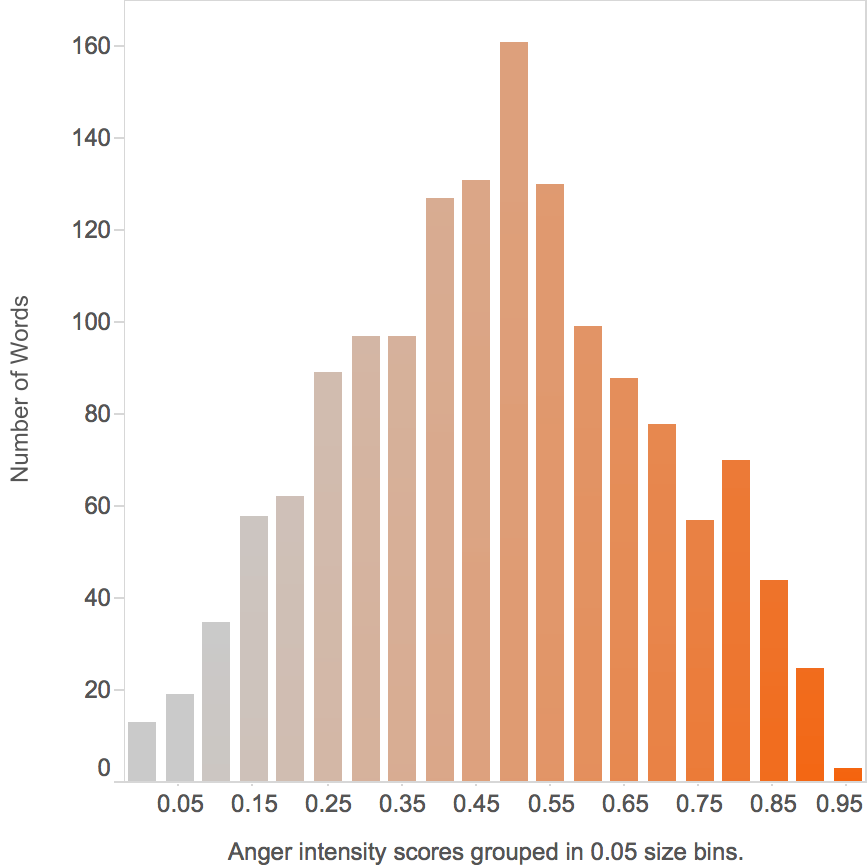}
\caption{A histogram of word--anger intensities. Anger intensity scores are grouped in bins of size 0.05. The colors of the bars go from gray to orange in increasing order of affect intensity.}
\label{fig:anger-histo}
\end{figure}

\noindent {\bf Annotation Aggregation:} The intensity scores were calculated from the BWS responses 
using a simple counting procedure \cite{Orme_2009,flynn2014}:
For each item, the score is the 
proportion of times the item was chosen as having the most intensity minus the 
proportion of times the item was chosen as having the least intensity.
The scores range from -1 (least emotion intensity) 
to 1 (the most emotion intensity).
Since degree of emotion is a unipolar scale, we linearly transform the -1 to 1 scores to scores in the range 0 (least emotion intensity) to 1 (the most emotion intensity).
We refer to the full list of words
along with their real-valued scores of affect intensity as the {\it NRC Affect Intensity Lexicon}.

\noindent  {\bf Distribution of Scores:} 
Figure \ref{fig:anger-histo} shows 
a histogram of word--anger intensities. The words are grouped into bins of scores 0--0.05, 0.05--0.1, and so on until 0.95--1.
Observe that the intensity scores have a normal distribution. The histograms for other emotions have a similar shape.

Table \ref{tab:ann} gives a summary of the number of items annotated and the number of annotations obtained.
Table \ref{tab:examples} shows some 
example entries from the lexicon.
The lexicon is made freely available. 


\section{Relationships of the Basic Emotions with Valence, Arousal, and Dominance}

\noindent  Even though the basic emotions model has long enjoyed the attention of psychologists, the valence--arousal--dominance (VAD) model \cite{russell2003core} is also widely accepted. 
According to the VAD model of affect, individual emotions are points in a three-dimensional space of valence (positiveness--negativeness), arousal (active--passive), and dominance (dominant--submissive). 
Both the basic emotions model and the VAD model 
offer different perspectives that help our understanding of emotions. However, there is little work relating the two models of emotion with each other. Much of the past work on textual utterances such as sentences and tweets, 
is based on exactly one or the other model (not both). 
For example, corpora annotated for emotions are either annotated only for the basic emotions \cite{SemEval2007,MohammadB17wassa} 
or only for valence, arousal, and dominance 
\cite{yu2016building,MohammadSK17,Nakov2016}.
\newcite{LREC18-TweetEmo} created the first dataset of tweets manually annotated for multiple affect dimensions from both the basic emotion model and the VAD model.
For each emotion dimension, they annotated the data for coarse classes (such as no anger, low anger, moderate anger, and high anger)
and also for fine real-valued scores indicating the intensity of emotion (anger, sadness, valence, etc.). They present an analysis of emotion intensities of tweets and their relationship with valence.

\begin{figure}[t]
\centering
\hspace*{-5mm} \includegraphics[width=2.3in]{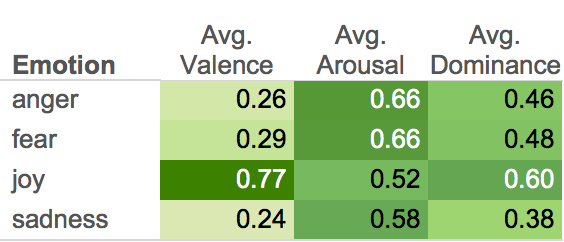}
\caption{Average valence, arousal, and dominance scores for each basic emotion. The cells are in shades of green with the darkness proportional to the score: lighter shades indicate low scores and darker shades indicate high scores.}
\label{fig:emo-vad-avg}
\end{figure}

Similar to the situation for textual corpora, {\it words} have been annotated largely either just for valence, arousal, and dominance ({\it ANEW} \cite{bradley1999affective}, the {\it Warriner Lexicon} \cite{warriner2013norms}, and the {\it NRC VAD
Lexicon} \cite{Mohammad-NRC-Norms}) or just for association with basic emotions (the NRC Emotion Lexicon \cite{MohammadT13,MohammadT10}).
Since all the words in the Affect Intensity Lexicon also have entries in the NRC VAD Lexicon \cite{Mohammad-NRC-Norms}, we now examine the relationship between 
the valence, arousal, and dominance scores across different basic emotions.

\begin{figure*}[t]
\centering
\hspace*{-5mm} \includegraphics[width=6.3in]{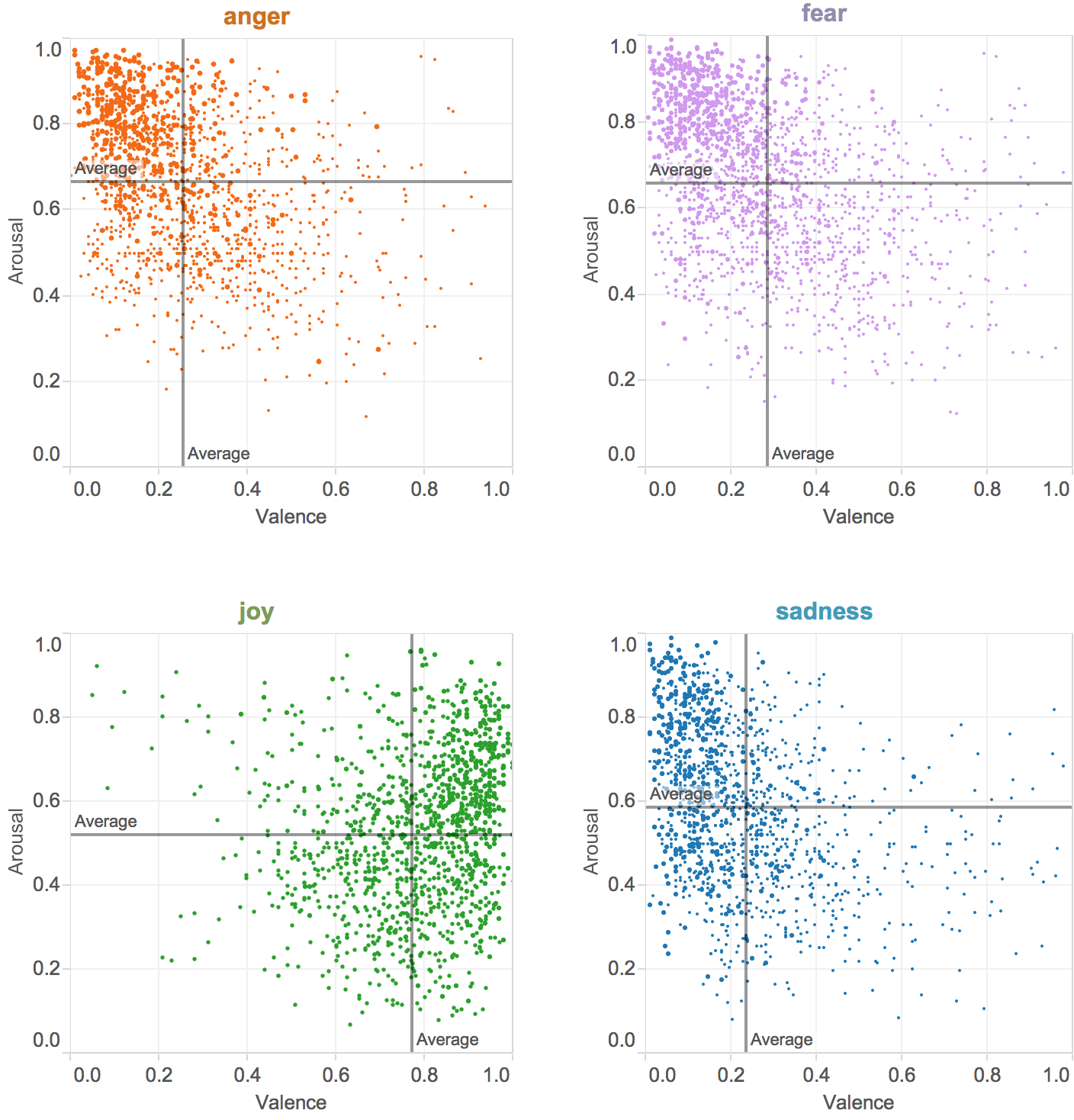}
\vspace*{-1mm}
\caption{Valence--Arousal scatter plots for words associated with each of the four basic emotions. For example, the anger plot on the top left has points for every anger word in the Affect Intensity Lexicon. The position of the point indicates its valence and arousal scores (as obtained from the NRC VAD Lexicon). The size of the point is proportional to the intensity of anger (as obtained from AIL). The size of the point is proportional to the intensity of the corresponding emotion. }
\vspace*{-4mm}
\label{fig:emo-VA}
\end{figure*}

\subsection{Valence, Arousal, and Dominance of Words in the Affect Intensity Lexicon}

\noindent  The NRC VAD Lexicon \cite{Mohammad-NRC-Norms} has valence, arousal, and dominance scores for over 20,000 commonly used English terms. It was created using best--worst scaling in a similar approach described earlier in this paper for obtaining emotion intensity scores. The three sets of scores range from 0 (lowest valence, arousal,  and dominance) to 1 (highest valence, arousal, and dominance). 

For each of the words in the Affect Intensity Lexicon, we looked up their entries in the NRC VAD Lexicon for scores of valence, arousal, and dominance.
Figure \ref{fig:emo-vad-avg} shows the average scores 
for each of the basic emotions. Figure \ref{fig:emo-VA} shows the scatter plot of the Affect Intensity Lexicon words across the orthogonal valence-arousal space. 

Observe that as expected, joy words have much higher valence scores (are much more positive) on average than the anger, fear, and sadness words. Joy words also have somewhat lower arousal scores (are more passive) on average than sadness words, which in turn have somewhat lower average arousal scores than anger and fear. 
Interestingly, anger and fear have a very similar profile of average VAD scores. Sadness words, on average, have the lowest average valence, followed by fear and anger.  

\begin{figure*}[t!]
\centering
\hspace*{-5mm} \includegraphics[width=5.4in]{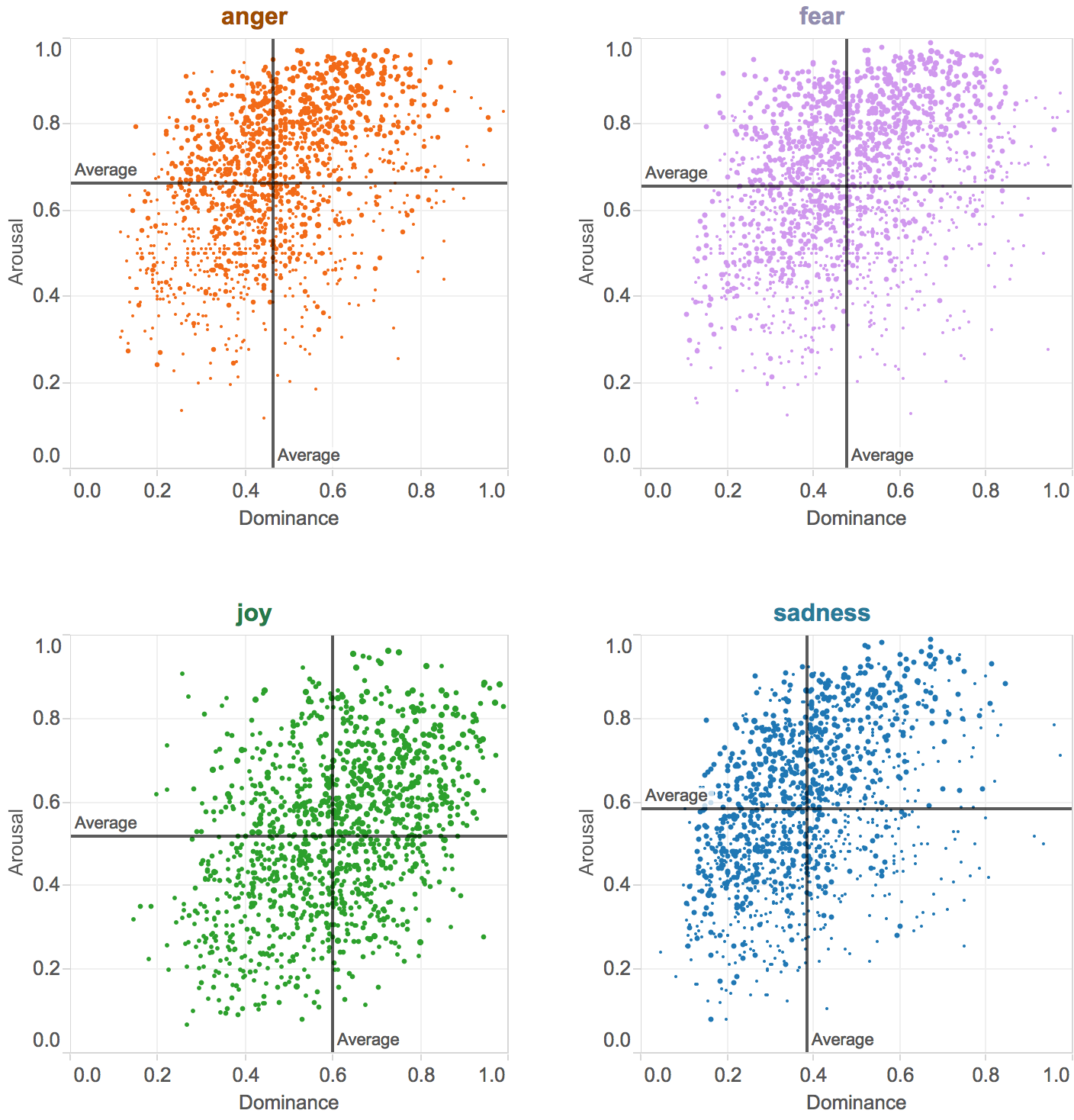}
\caption{Dominance--Arousal scatter plots for words associated with each of the four basic emotions.}
\label{fig:emo-DA}
\vspace*{2mm}
\end{figure*}

To determine whether the dominance--arousal space separates the three negative emotion words from each other, we generated the corresponding scatter plots as well.
See Figure \ref{fig:emo-DA}. Observe that words conveying negative emotions can belong to a wide and overlapping range of arousal and dominance scores. The range of scores now overlaps markedly with the joy words as well. 
Figure \ref{fig:emo-VD} in the Appendix shows the scatter plots for the valence--dominance space.

Overall, we observe that the  three negative emotions can be conveyed by words having a wide range of values for valence, arousal, and dominance.\footnote{The range is limited to the lower half of valence, but knowing valence is not sufficient to determine the precise basic emotion.}  
Let the words with emotion intensity scores greater than 0.5 be called the {\it upper-half subset}. The upper half subset includes words expressing medium to high emotion intensity.
Table \ref{tab:VADexamples} lists, for each emotion, the top four words that have highest and lowest valence, arousal, and dominance scores in the upper-half  subset of the emotion.
Note that for the negative emotions, the highest valence entries in the upper-half subset of anger are still expected to be somewhat negative.\footnote{For example, amongst the moderate-to-high anger terms, the highest valence term is still expected to be somewhat negative.}


\begin{table*}[t!]
\begin{center}
{\small
\begin{tabular}{lllllll}
\hline
Emotion &V$\uparrow\uparrow$	&V$\downarrow\downarrow$	&A$\uparrow\uparrow$	&A$\downarrow\downarrow$  & D$\uparrow\uparrow$	&D$\downarrow\downarrow$\\\hline
anger $> 0.5$  & {\it blaze  		} & {\it  shit		} & {\it  homicide		} & {\it  batter} & {\it  domination  	} & {\it  casualty} \\
 & {\it glare   	} & {\it  homicide 	} & {\it  terrorism		} & {\it  tiredofit	} & {\it  battle		} & {\it  idiots} \\
 & {\it incense 	} & {\it  murderous	} & {\it  violently		} & {\it  causality } & {\it  overbearing	} & {\it  slave} \\
 & {\it temper 		} & {\it  terrorist	} & {\it  enraged		} & {\it  cross		} & {\it  dictatorial	} & {\it  dishonest} \\[3pt]
fear $> 0.5$	 & {\it  seize				} & {\it  nightmare		} & {\it  abduction	} & {\it  senile	} & {\it  domination } & {\it  defenseless} \\
				 & {\it  meltdown			} & {\it  afraid		} & {\it  exorcism		} & {\it  coma		} & {\it  projectiles } & {\it  hopeless} \\
				 & {\it  retribution		} & {\it  homicide		} & {\it  homicide		} & {\it  stalk		} & {\it  dictator } & {\it  cowardly} \\
				 & {\it  enforce 			} & {\it  murderer		} & {\it  violently	} & {\it  hopeless	} & {\it  beastly } & {\it  casualty} \\[3pt]
joy $> 0.5$		 & {\it  generous	} & {\it  raving			} & {\it  elated	} & {\it  stressfree  	} & {\it  powerful } & {\it  silly} \\
				 & {\it  happily	} & {\it  zeal				} & {\it  excitation	} & {\it  peaceful	} & {\it  success  } & {\it  heheh} \\
				 & {\it  love		} & {\it  silly				} & {\it  euphoria	} & {\it  serenity 		} & {\it  triumphant } & {\it  weeeee} \\
				 & {\it  magnificent	} & {\it  boisterous	} & {\it  erotic	} & {\it  tranquility	} & {\it  winning } & {\it  snuggles} \\[3pt]
sadness $> 0.5$	 & {\it  meltdown	} & {\it  bankruptcy		} & {\it  abduction 	} & {\it  nothingness	} & {\it  warfare  		} & {\it  defenseless} \\
				 & {\it  console	} & {\it  disheartening		} & {\it  exorcism	} & {\it  alone				} & {\it  earthquake } & {\it  weakly} \\
				 & {\it  insurmountable	} & {\it  homicide	} & {\it  homicide	} & {\it  emptiness			} & {\it  bomber } & {\it  hopeless} \\
				 & {\it  longing	} & {\it  pain				} & {\it  terrorism	} & {\it   senile			} & {\it  unforgiving } & {\it  pity} \\ 
\hline
\end{tabular}
\vspace*{-1mm}
\caption{\label{tab:VADexamples} {The top four words that have highest and lowest valence (V), arousal (A), and dominance (D) scores, while also having an emotion intensity score greater than 0.5 (in the upper-half subset). The emotion intensity scores are taken from the NRC Affect Intensity Lexicon and valence, arousal, and dominance scores are taken from the NRC VAD Lexicon. $\uparrow\uparrow$ indicates the highest score entries. $\downarrow\downarrow$ indicates the lowest score entries.}}
}
\end{center}
\end{table*}

\section{Reliability of the Annotations}
\noindent  One cannot use standard inter-annotator agreement to determine quality of BWS annotations because
the disagreement that arises when a tuple has two items that are  close in emotion intensity is a
useful signal for BWS. For a given 4-tuple, if respondents are not able to consistently identify the word that  has
highest (or lowest) emotion intensity, then  the disagreement will lead to the two words obtaining
scores that are close to each other, which is the desired outcome. Thus a different measure of
quality of annotations must be utilized.

\begin{table}[t!]
\begin{center}
\begin{tabular}{lrrlrrlrrlr}
\hline
Emotion		&Spearman	&Pearson\\\hline
anger		&0.906		&0.912\\
fear		&0.910		&0.912\\
joy			&0.925		&0.924\\
sadness		&0.904		&0.909\\
\hline
\end{tabular}
\caption{\label{tab:shr} {Split-half reliabilities (as measured by Pearson correlation and Spearman rank correlation) for the anger, fear, joy, and sadness entries in the NRC Affect Intensity Lexicon.}}
\end{center}
\end{table}

A useful measure of quality is reproducibility of the end result---if repeated independent manual
annotations from multiple respondents result in similar intensity scores, then one can be confident
that the scores capture the true emotion intensities.  To assess this reproducibility, we calculate
average {\it split-half reliability (SHR)} over 100 trials. SHR is a commonly used approach to
determine consistency in psychological studies, that we employ as follows.   All annotations for an
item (in our case, tuples) are randomly split into two halves. Two sets of scores are produced
independently from the two halves.  Then the correlation between the two sets of scores is
calculated. If the annotations are of good quality, then the correlation between the two halves will
be high.  Table \ref{tab:shr} shows the split-half reliabilities for the anger, fear, joy, and sadness entries in the NRC Affect Intensity Lexicon. Observe that both the Pearson correlation and the Spearman rank correlations are above 0.9, indicating a high degree of
reproducibility. Note that SHR indicates the quality of annotations obtained when using only half the number of annotations; the correlations obtained when repeating the experiment  with four annotations for each 4-tuple is expected to be  higher than 0.91. Thus 0.91 is a lower bound on the quality of annotations obtained with four annotations per 4-tuple.

\section{Applications and Future Work}
\noindent  The  NRC Affect Intensity Lexicon  has many applications including automatic emotion analysis in a number of domains such as commerce, education, intelligence, and public health. 
The AIL was already used by several teams that participated  in the WASSA-2017 shared task on 
Emotion Intensity in Tweet \cite{MohammadB17wassa} as well as the SemEval-2018 Task 1: Affect in Tweets \cite{SemEval2018Task1} (including the teams that came first in both shared tasks). 
AIL is also useful in the building of natural language generation systems. 

We are currently using the NRC Affect Intensity Lexicon along with tweets datasets that were annotated for emotion intensity ({\it Tweet Emotion Intensity Dataset} \cite{MohammadB17starsem}), to test the extent to which people convey strong emotions in tweets using high-intensity emotion words. We will also use the lexicon to identify syllables that consistently tend to occur in words with strong affect associations. This has implications in understanding how some syllables and sounds have a tendency to occur in words referring to semantically related concepts. Identifying emotions associated with a syllable is also useful in generating names for literary characters and commercial products.  

The lexicon also has applications in the areas of digital humanities and literary analysis, where it can be used to identify high-intensity words. The NRC  Affect Intensity Lexicon can also be used as a source of gold intensity scores to evaluate automatic methods of determining word affect intensity.

\section{Conclusions}
\noindent  This paper describes how we created the {\it NRC Affect Intensity Lexicon}---a crowdsourced lexicon that captures word--affect intensities for four basic emotions: anger, fear, joy, and sadness. We used a technique called best--worst scaling (BWS) to obtain fine-grained scores (and word rankings). BWS addresses issues of annotation consistency that plague traditional rating scale methods of annotation.  
We show that repeat annotations of the terms in the Affect Intensity Lexicon with independent annotators lead to affect association scores that are close to the scores obtained originally (split-half reliability:  $rho = 0.92$, $r = 0.91$).
The fine-grained scores obtained with BWS  and the high correlations on repeat annotations indicate that BWS is both markedly discriminative (helps identify small differences in affect intensity) and markedly reliable (provides stable outcomes). 

The Affect Intensity Lexicon has applications in automatic emotion analysis as well as in understanding  affect composition---how affect  of a sentence is impacted by the affect of its constituent words. We will continue to add entries for other emotions such as disgust,  trust, surprise, and anticipation. We will use the lexicon to study the role emotion words play in high emotion intensity tweets, using the Tweet Emotion Intensity Dataset that has intensity scores for whole tweets. We will also use the lexicon to determine syllables and phonetic sounds that are associated with particular affect categories, that is, syllables that tend to occur more often than average in affect-associated words. The lexicon is made freely available. 

\section*{Acknowledgments}

\noindent Many thanks to Svetlana Kiritchenko and Tara Small for  helpful discussions.

\newpage

\begin{figure*}[t!]
\centering
\hspace*{-5mm} \includegraphics[width=6.3in]{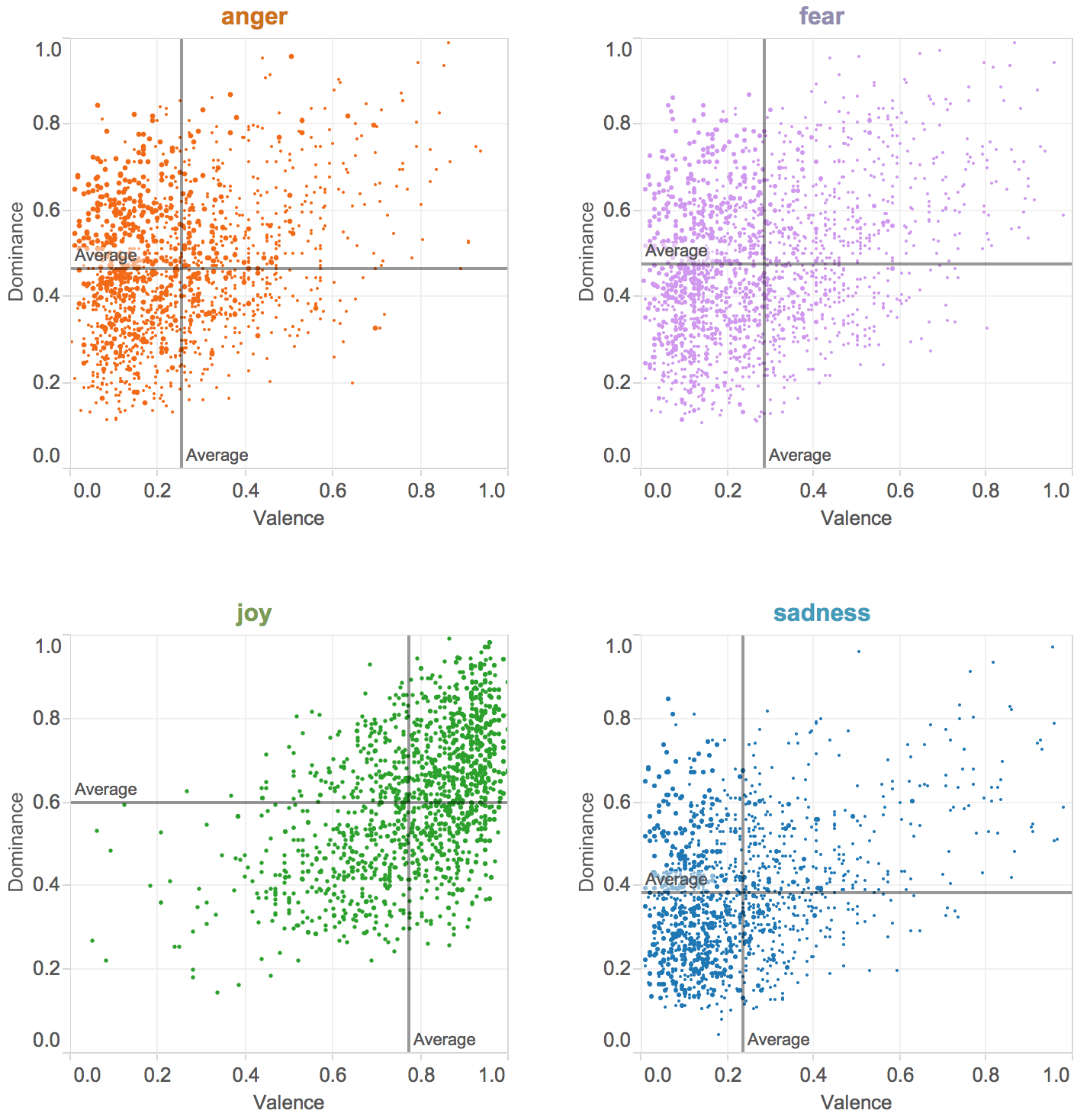}
\caption{Valence--Dominance scatter plots for words associated with each of the four basic emotions. The size of the point is proportional to the intensity of the corresponding emotion. For example, the anger plot on the top left has points for every anger word in the Affect Intensity Lexicon. The position of the point indicates its valence and dominance scores (as obtained from the NRC VAD Lexicon). The size of the point is proportional to the intensity of anger (as obtained from the Affect Intensity lexicon).}
\label{fig:emo-VD}
\end{figure*}
\section{Bibliographical References}
\bibliography{maxdiff}
\bibliographystyle{lrec}


\section{Appendix}

\noindent Figure \ref{fig:emo-VD} 
shows scatter plots for the four basic emotions in the valence-dominance space.


\end{document}